\documentclass[a4paper,10pt,bibtotoc]{scrartcl}
\usepackage{a4wide}
\usepackage{graphicx}        % For eps figures, newer & more powerfull
\usepackage{color}           % For color text: \color command
\usepackage{breakurl}        % For breaking URLs easily trough lines
\usepackage{amsmath}
\usepackage{breqn}
\usepackage{bbold}
\graphicspath{{figures/}}
\usepackage{subfig}
\usepackage{graphicx,wrapfig,lipsum}
\usepackage{placeins}
\DeclareMathAlphabet{\mymathbb}{U}{bbold}{m}{n}

\usepackage{algorithm}
\usepackage[noend]{algpseudocode}

\usepackage{titling}

\makeatletter
\def\BState{\State\hskip-\ALG@thistlm}
\makeatother

\newcount\colveccount
\newcommand*\colvec[1]{
        \global\colveccount#1
        \begin{pmatrix}
        \colvecnext
}
\def\colvecnext#1{
        #1
        \global\advance\colveccount-1
        \ifnum\colveccount>0
                \\
                \expandafter\colvecnext
        \else
                \end{pmatrix}
        \fi
}

\title{Implicit recurrent networks: \\ A novel approach to stationary input processing with recurrent neural networks in deep learning.}

\author
{Sebastian Sanokowski,$^{1\ast}$\\
\\
\normalsize{$^{1}$Friedrich-Alexander Universität Erlangen-Nürnberg; Department of Physics,}\\
\normalsize{$^\ast$ E-mail:  sebastian.sanokowski@fau.de}
}

\begin{document}

\maketitle

%\author[addressref={aff1},corref,email={sebastiansanokowski@gmail.com}]{\inits{S.}\fnm{Sebastian}~\lnm{Sanokowski}}
%\author[addressref=aff1,email={claus.metzner@gmail.com}]{\inits{C.}\fnm{Claus}~\lnm{Metzner}}
%\author[addressref=aff2,email={Patrick.Krauss@uk-erlangen.de}]{\inits{P.}\fnm{Patrick}~\lnm{Krauss}}
%\author{\inits{}\fnm{}~\lnm{}\orcid{}}
%\author{P.~\surname{Author-a}$^{1}$\sep
%        E.~\surname{Author-b}$^{1}$\sep
%        M.~\surname{Author-c}$^{2}$      
%       }

%   \institute{$^{1}$ First affiliation
%                     email: \url{e.mail-a} email: \url{e.mail-b}\\ 
%              $^{2}$ Second affiliation
%                     email: \url{e.mail-c} \\
%             }
%\address[id=aff1]{Friedrich-Alexander Universität Erlangen-Nürnberg; Department of Physics}
%\address[id=aff2]{Universitätsklinikum Erlangen}

%\runningauthor{Sanokowski et al.}
%\runningtitle{Implicit recurrent networks}

\begin{abstract}

The brain cortex, which processes visual, auditory and sensory data in the brain, is known to have many recurrent connections within its layers and from higher to lower layers.
But, in the case of machine learning with neural networks, it is generally assumed that strict feed-forward architectures are suitable for static input data, such as images, whereas recurrent networks are required mainly for the processing of sequential input, such as language. However, it is not clear whether also processing of static input data benefits from recurrent connectivity. 
In this work, we introduce and test a novel implementation of recurrent neural networks with lateral and feed-back connections into deep learning.
This departure from the strict feed-forward structure prevents the use of the standard error backpropagation algorithm for training the networks. Therefore we provide an algorithm which implements the backpropagation algorithm on a implicit implementation of recurrent networks, which is different from state-of-the-art implementations of recurrent neural networks. Our method, in contrast to current recurrent neural networks, eliminates the use of long chains of derivatives due to many iterative update steps, which makes learning computationally less costly.
It turns out that the presence of recurrent intra-layer connections within a one-layer implicit recurrent network enhances the performance of neural networks considerably: 
A single-layer implicit recurrent network is able to solve the XOR problem, while a feed-forward network with monotonically increasing activation function fails at this task. Finally, we demonstrate that a two-layer implicit recurrent architecture leads to a better performance in a regression task of physical parameters from the measured trajectory of a damped pendulum.
\end{abstract}
%\keywords{Artificial neural networks, Recurrent neural networks, Deep learning, Implicit recurrent networks}

%-------------------------------------------------
\thispagestyle{empty}
\newpage
\setcounter{page}{1}
\section{Introduction}

There is some research, which indicates that human vision for stationary input processing is not merely a feed forward process \cite{kar2019evidence, kietzmann2019recurrence, johnson2005recognition}.
Therefore the question arises, whether deep learning may also benefit from recurrent connections when stationary inputs, such as images, are processed.
In order to make it plausible, that recurrence does play a role in human vision, we will take the example of the famous illusion ''My Wife and My Mother-in-Law'', where one can either see a young woman or an old woman in this drawing.
It is well known that neurons within the brain or within artificial neural networks act as feature detectors, where the detected features in higher layers are more complex than in lower layers.
Therefore, if our brain detects a young woman in this picture, there may be active neurons within lower brain regions, which recognise a necklace or the chin of the young woman.
But, if one recognises an old woman, the active features within lower brain regions change and instead other neurons, which may instead detect the nose and the mouth of the old woman, become active. This example illustrates that the final prediction of the human brain influences neural activation in lower brain regions, which can only be explained with recurrent connections.
Therefore, by using such recurrent mechanisms in artificial neural networks error correction from higher layers into lower ones may be possible.\\
In a feed-forward process all of these lower features would have to be active, independent of what the network recognises in the end. While this may be not a problem if this network is merely used for classification, one could assume that incorrectly detected features are a problem in architectures, where lower layers are shared or where lower features are used for some other task.
Another advantage of recurrent connections in deep learning may be that, for example,  lateral recurrence within a layer would bring some additional interaction within a layer. 
This may lead to advantages during the training process, where connected neurons may recognise more easily that they are learning similar features and could therefore learn different tasks at an earlier stage during the learning phase.
Furthermore, the additional interaction due to recurrence may lead to more computational power of the network. This assumption will be studied in this paper.

\section{Results}

\subsection{Problems with state-of-the-art recurrent neural networks for stationary input processing}

In current state-of-the-art models for recurrent neural networks for stationary input processing, such as images, so called iterative recurrent networks are used.
In iterative recurrent networks, like they have been used in O'Reilly et al. \cite{o2013recurrent} and in M. Ernst et al. \cite{ernst2019recurrent}, the recurrent neural activation is updated iteratively over a few time steps.

In order to explain the principle of iterative recurrent networks, we are going to consider the simplest case of a single iterative recurrent neuron $Y$. This neuron receives a stationary input value $X$, which is connected via the weight Q to neuron $Y$. Additionally, $Y$ is recurrently connected to itself through the weight $W$.

The activation of this neuron is then given by
\begin{equation*}
    Y(t) = f(W \cdot Y(t-1) + Q \cdot X + T),
\end{equation*}
where $T$ is the bias of this neuron and $f(\cdot)$ is the sigmoid activation function.
The activation of this neuron is then calculated iteratively by starting at $Y(t=0) = 0$ and calculating the consecutive time steps from there on up to $Y(t=n)$.\\
Here, we identify the first problem of this method, where it is not clear how many iterative updates should be used. 
The next problem occurs, when we want to train these recurrent networks with backpropagation.
In order to apply gradient descent, it would for example be necessary to calculate the derivative with respect to $W$.
In our simple example, we obtain 
\begin{equation*}
    \frac{\partial Y(t)}{\partial W} = f^{\prime}(\widetilde{X}) \cdot (W \cdot \frac{\partial Y(t - 1)}{\partial W} + Y(t-1)),
\end{equation*}
where $\frac{\partial Y(t - 1)}{\partial W}$ again depends on $Y(t-2)$ which also depends on $W$. 
Therefore, in order to obtain the final derivative, the chain rule has to be applied $n$ times up to $Y(t=0)$.
Hence, for every iterative time step which is used in this network the derivatives become more complicated and if too many time steps are used, this calculation becomes intractable.
That is why in iterative recurrent networks usually only a few iterative updates are used.

However, these iterative networks have an equilibrium, when $\vec{Y}(t) = \vec{Y}(t-1)$, which is often reached after a few iterative updates of the network. But, if the number of neurons within a layer is too large, iterative updates will not converge to this equilibrium.

\subsection{Intuition of implicit recurrent networks}

Instead of using iterative recurrent networks, we use the alternative approach of implicit activation functions with which the equilibrium state can be calculated immediately.

In the simplest case of a single recurrent neuron, the implicit activation function is given by
\begin{equation}
    Y = f(W \cdot Y + Q \cdot X + T).
    \label{eq:1-neuron-implicit}
\end{equation}

Here, in order to obtain the activation of the neuron the equilibrium $Y$ has to be found, where the left and the right side of the equation equal one another.

Next, we will show that this mathematical formulation will make it possible to calculate derivatives directly at the equilibrium state of recurrent neural networks. These derivatives are, in contrast to derivatives of iterative recurrent networks, not computationally expensive.

\begin{figure}
    \centering
    \includegraphics[width = 0.6\textwidth]{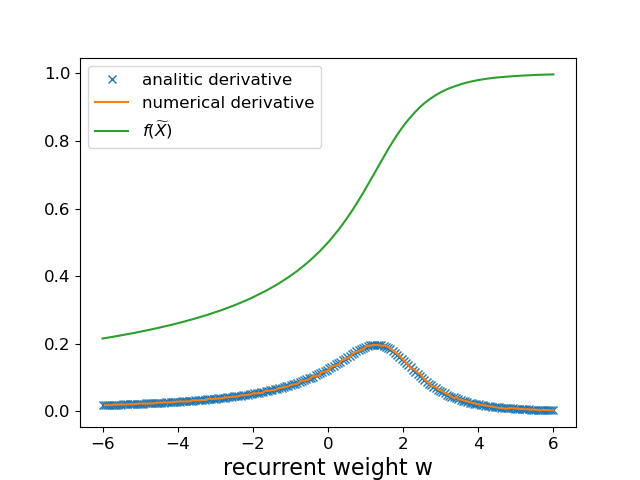}
    \caption{Comparison of the numerical derivative and the derivative from equation \ref{eq:1_implicit_neuron}. The numerical derivative fits perfectly to the derivative which we have derived. The numerical derivative is calculated by calculating the distance between two neighbouring points of $f(\widetilde{X})$ and dividing it by their distance.}
    \label{fig:num_der}
\end{figure}

Now, if we calculate the derivative with respect to $W$, we obtain 
\begin{equation*}
    \frac{\partial Y}{\partial W} = f^{\prime}(\widetilde{X}) \cdot (W \cdot \frac{\partial Y}{\partial W} + Y),
\end{equation*}
where $\widetilde{X} = W \cdot Y + Q \cdot X + T$ is the overall input of this neuron.
Here, the derivative $\frac{\partial Y}{\partial W}$ appears on the right side and on the left side of the equation.
Therefore, we obtain a linear equation which has to be solved for $\frac{\partial Y}{\partial W}$, in order to obtain the final derivative.
After solving this equation one obtains
\begin{equation}
    \frac{\partial Y}{\partial W} = \frac{f^{\prime}(\widetilde{X}) \cdot Y}{ 1 - f^{\prime}(\widetilde{X}) \cdot W},
    \label{eq:1_implicit_neuron}
\end{equation}

which fits perfectly to numerical approximations of this derivative, as can be seen in figure \ref{fig:num_der}.
Thus, by changing the mathematical formulation of iterative recurrent networks to implicit recurrent networks, it is possible to calculate the derivatives of this neuron in a way which is computationally not expensive.

Such implicit recurrent networks were originally introduced by P. Földiák \cite{foldiak1990forming}, where they have been used in so called Hebbian neural networks, which are trained with local Hebbian and Anti-Hebbian learning rules.
It is well known how to train these networks with Hebbian- and Anti-Hebbian learning rules, as it is shown in \cite{gerhard2009robust} or \cite{foldiak1990forming}, but to our knowledge nobody has formulated a way on how to train these networks with backpropagation.
Therefore, in this paper, we find a mathematical formulation for a one-layer implicit recurrent neural network in subsection \ref{subsec:one-layer}, which has lateral connections within the layer. Furthermore, in subsection \ref{subsec:two-layer} we write down a formulation for a two-layer fully recurrent implicit network which has additional feed-back connections from higher to lower layers.

\subsection{Tests on the XOR-Problem}

\begin{figure}
    \centering
    \subfloat[Network architecture]{\includegraphics[width =0.4\textwidth]{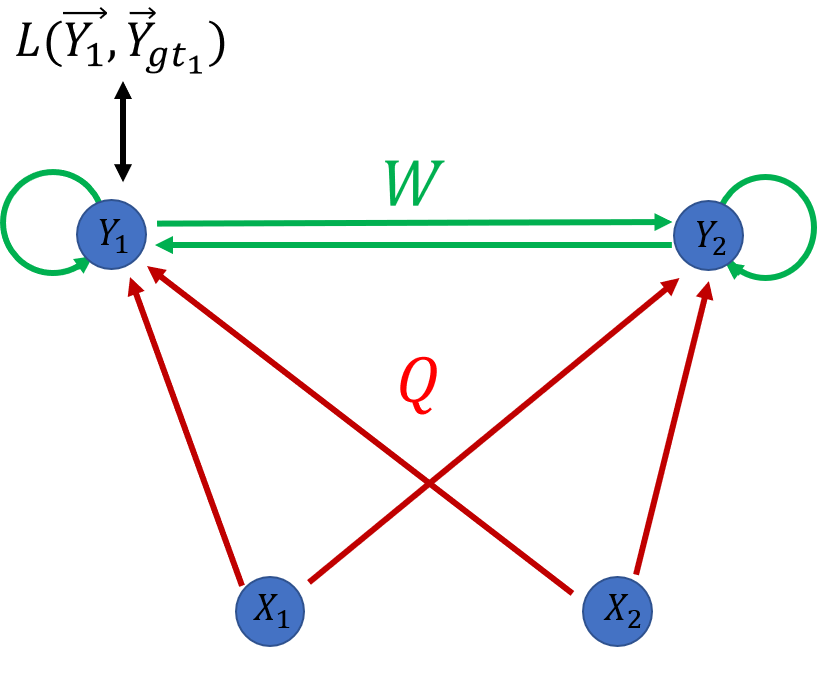}}
    \subfloat[Error over epochs]{\includegraphics[width = .4\textwidth]{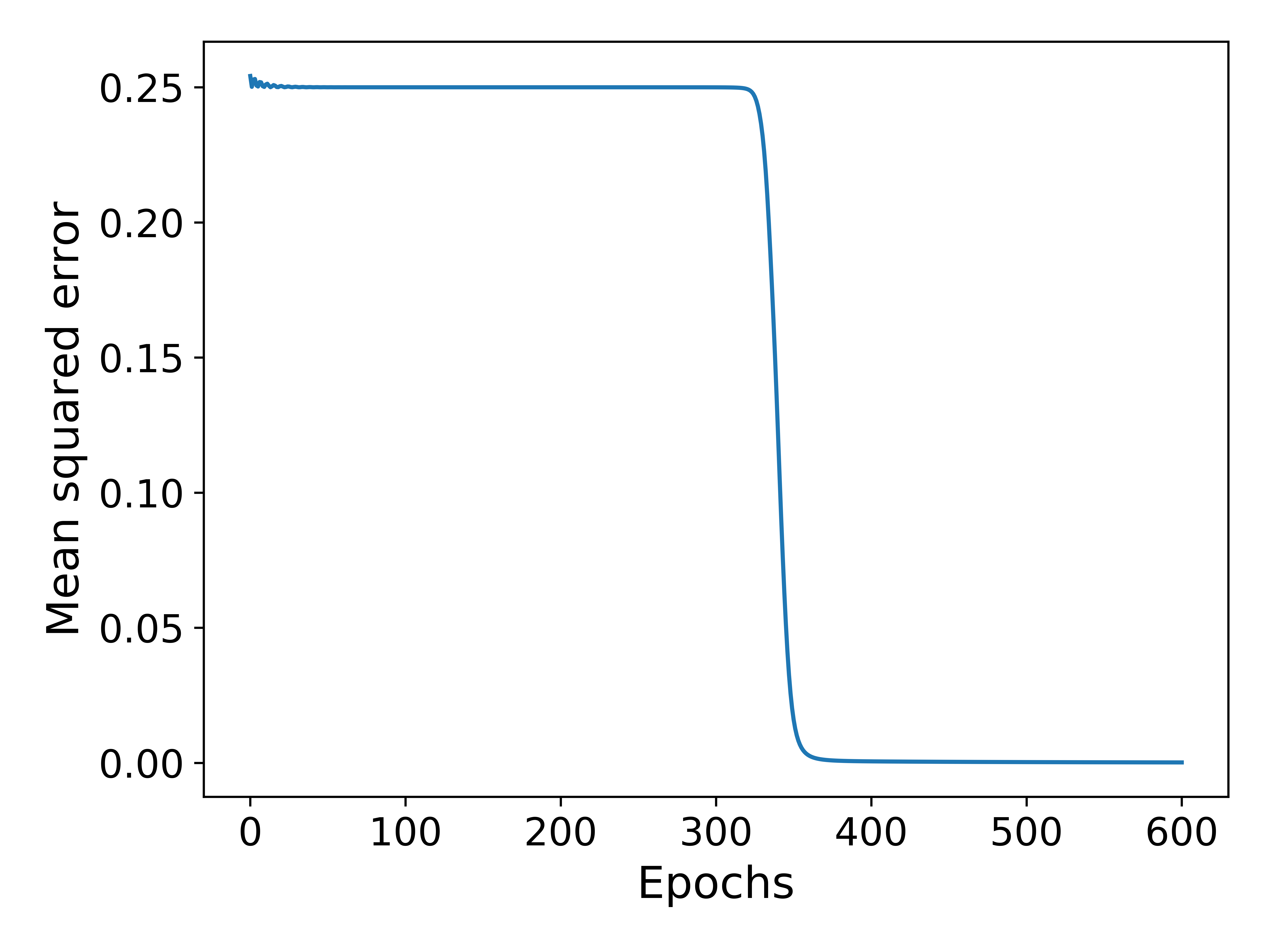}} 

    \centering
    \caption{Training results of a one-layer implicit recurrent network with two neurons. One Neuron $Y_1$ is trained directly on the XOR-Problem. Neuron $Y_2$ is trained indirectly due to recurrence. As can be seen in the error plot over the epochs in sub-figure (b), this network is able to solve this problem with a very low error rate.}
    \label{fig:XOR-implicit}
\end{figure}

\begin{figure}
\centering
\includegraphics[width=0.35\textwidth]{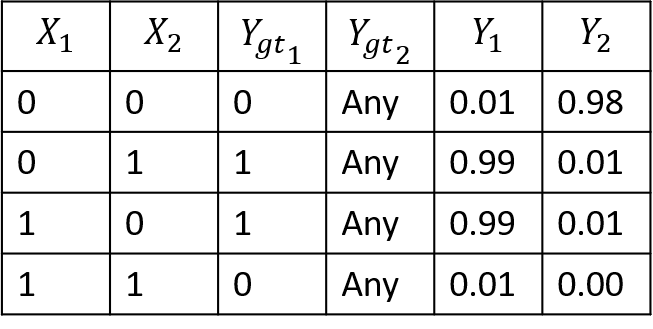}
\caption{Tabular with input data, the corresponding ground truth data and the output data after training for the simulation of the one-layer implicit recurrent network on the XOR-Problem. Only the output of neuron $Y_1$ is minimized through the loss function to the ground truth data $Y_{gt_1}$. Here, neuron $Y_2$ is allowed to generate any output. As can be seen neuron $Y_1$ learns an output, which is very close to the ground truth of the XOR-Problem.}
\label{tab:XOR-implicit}
\end{figure}

It is well known that a single feed-forward neuron, which has a monotonically increasing activation function can not solve the XOR-Problem.
Therefore, since a single feed-forward neuron can not solve this problem, a one-layer feed forward network also can not solve this problem because in feed-forward networks there are not interactions between neurons of the same layer.
But, in contrast to a feed-forward network a one-layer implicit recurrent network does have interactions within its layer.
Therefore, we try to solve the XOR-Problem with a one-layer implicit recurrent neural network.
Here, as depicted in figure \ref{fig:XOR-implicit}, we use two neurons and train only neuron $Y_1$ directly on the XOR-Problem through the minimisation of the mean squared error loss function.
The other neuron $Y_2$ is trained indirectly due to recurrence. Therefore, the other neuron $Y_2$ simply learns an output, which supports $Y_1$ in solving the XOR-Problem.
Results of this simulation are depicted in figure \ref{fig:XOR-implicit} and in tabular \ref{tab:XOR-implicit}, where we can see that this network is able to solve the XOR Problem with a very low error.
As can be seen in tabular \ref{tab:XOR-implicit}, the second neuron $Y_2$ learns the output of a NOR-gate, which is a subgate which supports $Y_1$ in solving the XOR-Problem.

\begin{figure}
    \centering
    \subfloat[Network architecture]{\includegraphics[width =0.4\textwidth]{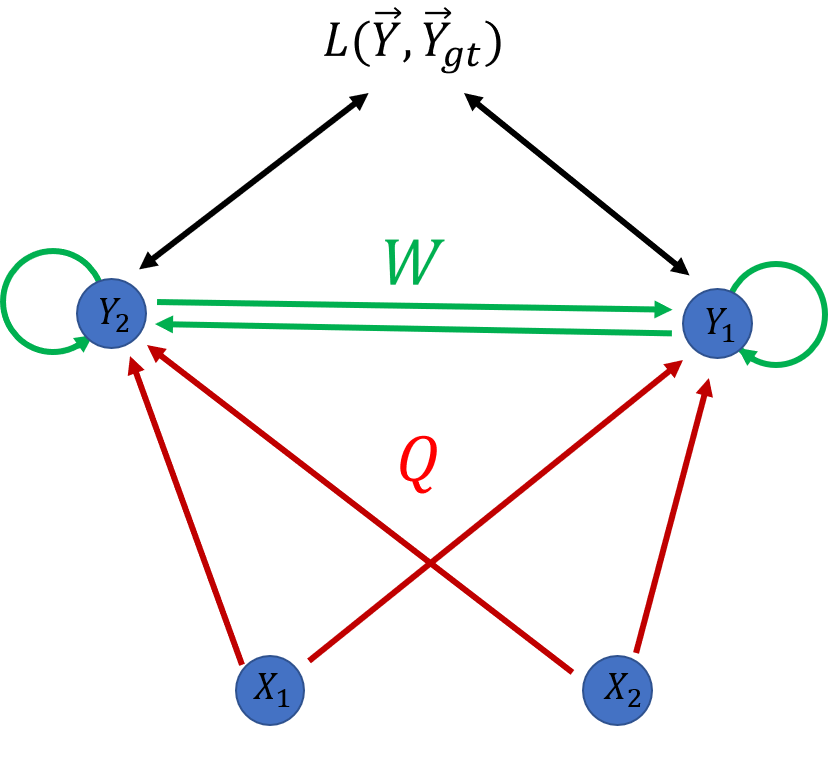}}
    \subfloat[Error over epochs]{\includegraphics[width = .4\textwidth]{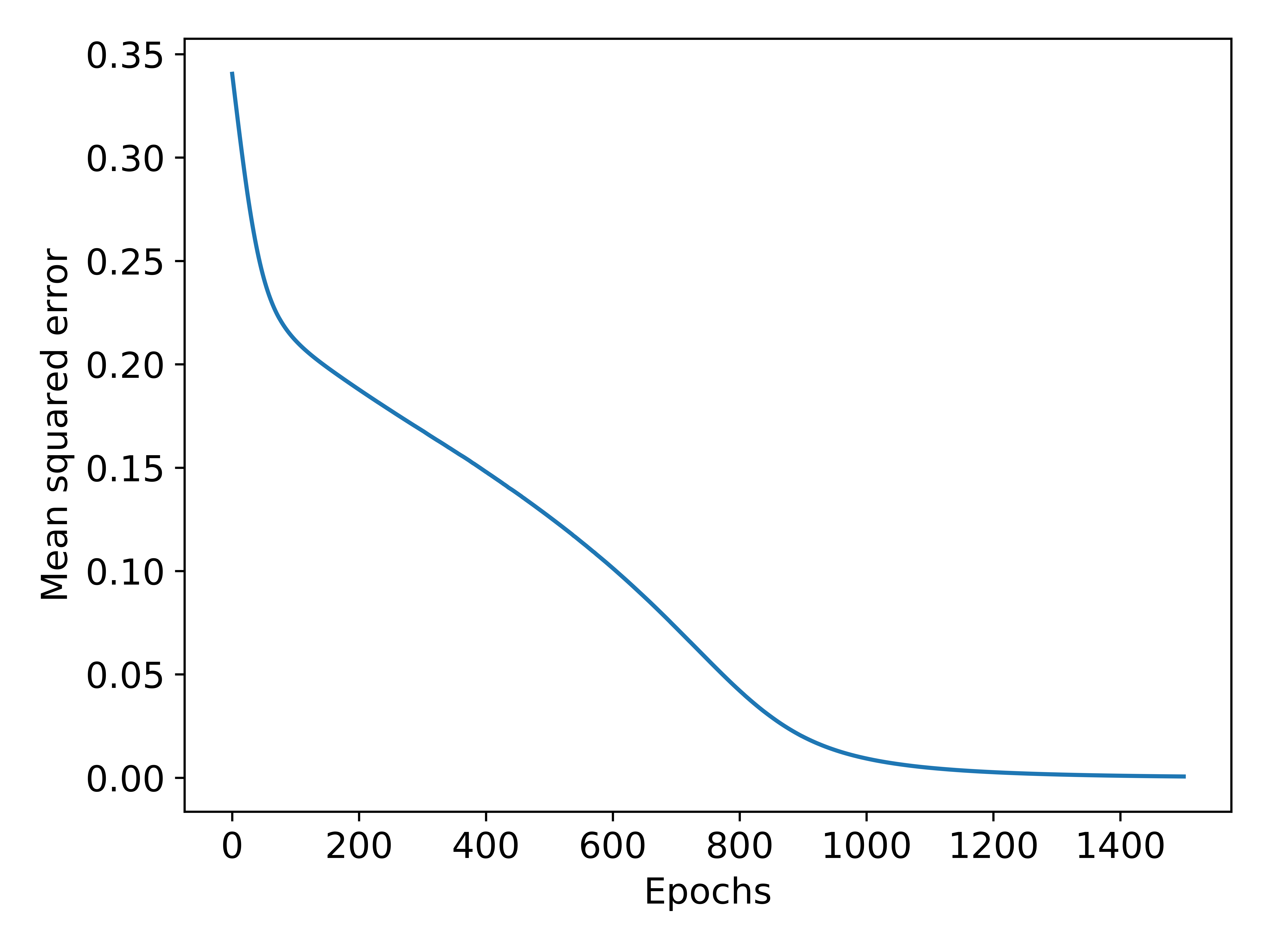}} 

    \centering
    \caption{Training results of a semi-gradient one-layer implicit recurrent network with two neurons. One Neuron $Y_1$ is trained directly on the XOR-Problem and neuron $Y_2$ is trained to a sub-problem. As can be seen in the error plot over the epochs in sub-figure (b), this network is able to solve this problem with a very low error rate.}
    \label{fig:XOR-semi-gradient-implicit}
\end{figure}

\begin{figure}
\centering
\includegraphics[width=0.35\textwidth]{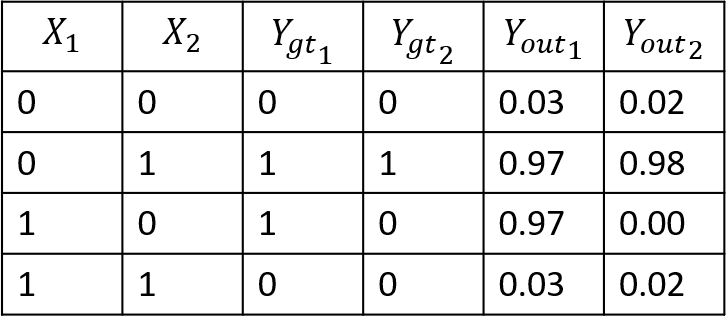}
\caption{Tabular with input data, the corresponding ground truth data and the output data after training for the simulation of the semi-gradient one-layer implicit recurrent network on the XOR-Problem. Here, $Y_1$ is trained directly to the XOR-Problem and $Y_2$ is trained to a simpler logic operation. As can be seen, even with the semi-gradient method neuron $Y_1$ learns an output, which is very close to the ground truth of the XOR-Problem. }
\label{tab:XOR-semi-gradient-implicit}
\end{figure}

To simplify, we also introduce a semi-gradient method for training such recurrent neural networks, as they are explained in the methods section in \ref{subsec:semi-gradient}.
Due to the approximation, this system loses the ability of indirect training of neurons within the same layer. Therefore, as can be seen in figure \ref{fig:XOR-semi-gradient-implicit} and in tabular \ref{tab:XOR-semi-gradient-implicit} the output of both neurons have to be trained to a ground truth target.
But still, as can be seen on the mean squared error in figure \ref{fig:XOR-semi-gradient-implicit} (b) or on the output in tabular \ref{tab:XOR-semi-gradient-implicit}, the network is still able to solve the XOR-Problem.
Thus, despite of the approximation, the network still remains computationally stronger than state-of-the-art feed forward networks.

\subsection{Performance on a regression problem in physics}

Next, we train two-layer feed-forward networks, two-layer implicit recurrent networks and semi-gradient two-layer implicit recurrent networks and compare their performance to one another on a regression problem.
Here, we generate the trajectory of a damped oscillator $X(t)$, which can be described by the differential equation 

\begin{equation*}
    \ddot{x} + 2 \cdot \delta \cdot \dot{x} + \omega_{0}^2 \cdot x = 0,
\end{equation*}

where $\omega_0$ and $\delta$ are the only parameters that govern this system.

In this problem, we generate for a given pair of $\omega_0$ and $\delta$ the trajectory of this system, which is then put into the network. The network is then supposed to reconstruct $\omega_0$ and $\delta$.
In our simulations, we discretize the trajectory with a time step of $\Delta t = 0.1$ and generated the dynamics with random starting positions $x_0$ and starting velocities $v_0$, which were drawn from a normal distribution $N(0,2)$. Additionally, $\omega_0$ is drawn form $[1,2]$ and $\delta$ is drawn form $[0,2]$. We have then separated the dataset with $4 \cdot 10^6$ samples into a training- and test dataset with the ratio of $4:1$. It is made sure, that every combination of $x_0, v_0, \omega_0$ and $\delta$ occurs only once in the whole dataset.
In our simulations we perform for every epoch 200 gradient steps over six batches. Every four epochs we calculate the mean squared error on the whole test dataset. All simulations are repeated 12 times over 400 epochs.
 
\begin{figure}
    \centering
    \subfloat[Comparison of the train error]{\includegraphics[width =0.5\textwidth]{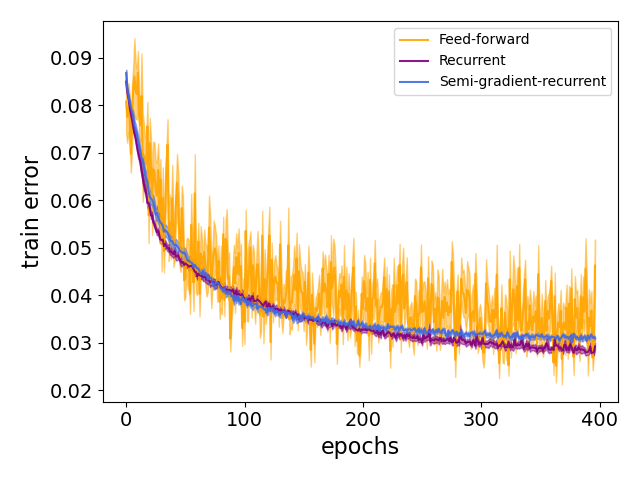}}
    \subfloat[Comparison of the test error]{\includegraphics[width = .5\textwidth]{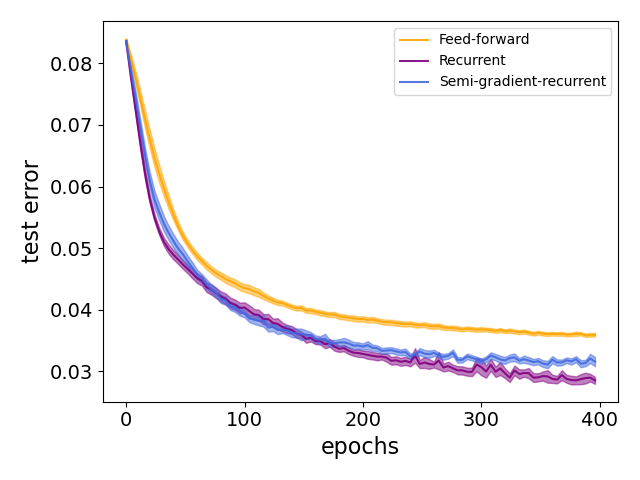}} 

    \centering
    \caption{Comparison of the performance of a two-layer feed-forward network, a semi-gradient two-layer implicit recurrent network and a semi-gradient two-layer implicit recurrent network on the regression problem of a damped pendulum. Results are averaged over 12 runs. The shaded region indicates the standard deviation of the mean.}
    \label{fig:regressehn_nh_5}
\end{figure}

The results of a simulation on this problem with a networks with $5$ hidden neurons are shown in figure \ref{fig:regressehn_nh_5}.
In figure \ref{fig:regressehn_nh_5} (a), we can see the mean train error performance of each network and the shaded region indicates with the standard deviation of the mean.
Here, the two-layer implicit recurrent network has the best performance, followed by the semi-gradient two-layer recurrent network.
The feed-forward network has the worst performance and interestingly during training it has much higher variance than its recurrent counterparts. This may indicate that the prediction of implicit recurrent networks is more consistent.

In figure \ref{fig:regressehn_nh_5} (b), the test error of each network is plotted over the epochs, where again the two-layer implicit recurrent network shows the best performance.

\begin{figure}
    \centering
    \subfloat[Final test error over number of hidden neurons]{\includegraphics[width =0.5\textwidth]{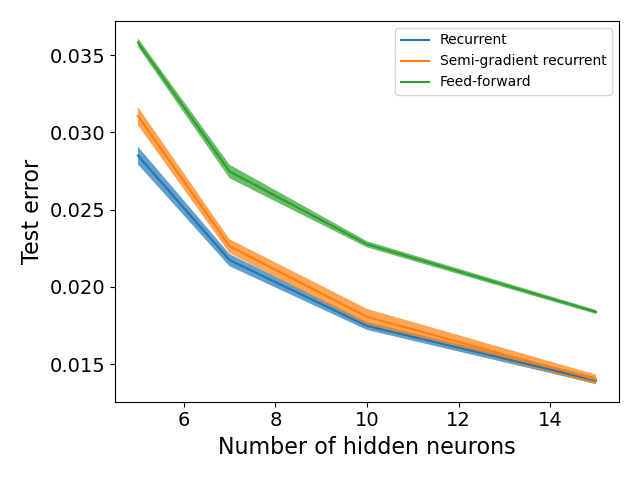}}
    \subfloat[Final test error over number of parameters]{\includegraphics[width =.5\textwidth]{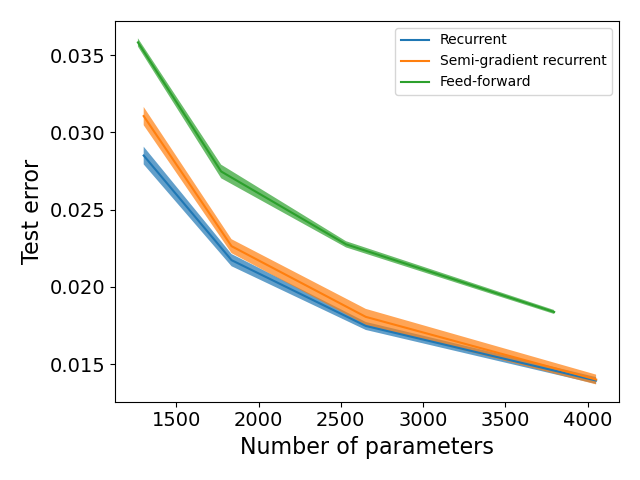}} 

    \centering
    \caption{Comparison of the final test error of a two-layer feed-forward network, a semi-gradient two-layer implicit recurrent network and a semi-gradient two-layer implicit recurrent network on the regression problem of a damped pendulum. In (a) the final test error is plotted over the number of hidden neurons of the trained networks and in (b) the final test error is plotted over the number of parameters. Results are averaged over 12 runs. The shaded region indicates the standard deviation of the mean. }
    \label{fig:regression_different_nhs}
\end{figure}

Additionally, we study the dependence of the final test error on the number of hidden neurons and on the number of parameters of each network architecture. Here, we train each network for a given number of hidden neurons over 400 epochs and plot the final test error in figure \ref{fig:regression_different_nhs} (a). In figure \ref{fig:regression_different_nhs} we take the results from \ref{fig:regression_different_nhs} (a), but plot the final test error over the number of parameters. 
Here, we see that two-layer implicit recurrent networks perform much better than their two-layer feed-forward counterparts. If we compare two-layer implicit recurrent networks to their semi-gradient counterpart, we see that for a small number of hidden neurons the semi-gradient method performs significantly worse. But, as the number of hidden neurons increases the disadvantage of semi-gradient implicit recurrent networks becomes smaller.
This may suggest, that if many neurons are used it is sufficient to use semi-gradient method instead of the using the exact derivatives. Though, we think this has to be studied more rigorously.

\FloatBarrier
 
\section{Methods} %%%%%%%%%%%%%%%%%%%%%%%%%%%%%%%%%%%%%%%%
      \label{S-general}      

\subsection{One-layer implicit recurrent networks}
\label{subsec:one-layer}

A one-layer implicit recurrent network receives an input $\vec{X}$ which is connected via a weight matrix $Q$ into a layer $\vec{Y}$, which is recurrently connected to itself via recurrent weights $W$.

The activation function of this layer is then given by 

\begin{equation}
    \vec{Y} = f(W \cdot \vec{Y} + Q \cdot \vec{X} + \vec{T}),
    \label{eq:impl_function_methods}
\end{equation}

where $\vec{Y}$ is solved with the method described in the method section \ref{subsec:equi}.

The derivatives with respect to the weights of $W$, which have to be used in order to perform gradient descend are given by 

\begin{equation}
     J_W = ( \mathbb{1} - f^{\prime} \odot W)^{-1} f^{\prime} \odot \delta_{\vec{Y}},
     \label{eq:dw_methods}
\end{equation}

where $(J_W)_{ijm} :=  \frac{\partial Y_i}{\partial W_{jm}}$, $(\delta_{\vec{Y}})_{ijm} := \delta_{ij}  \cdot Y_m $ and $(f^{\prime})_{ij} := \frac{\partial f(\widetilde{X}_i)}{\partial \widetilde{X}_i}$.
Here, $\odot$ is the element wise multiplication sign.

The derivatives with respect to the recurrent weights of $Q$ are given by
\begin{equation}
     J_Q = ( \mathbb{1} - f^{\prime} \odot W)^{-1} f^{\prime} \odot \delta_{\vec{X}},
     \label{eq:dq_methods}
\end{equation}

where $(J_Q)_{ijm} = \frac{\partial Y_i}{\partial Q_{jm}}$ and $(\delta_{\vec{X}})_{ijm} = \delta_{ij}  \cdot X_m$.

And finally, the derivatives with respect to the biases $T$ are given by
\begin{equation}
     J_T = ( \mathbb{1} - f^{\prime} \odot W)^{-1} f^{\prime} \odot  \mathbb{1},
     \label{eq:dt_methods}
\end{equation}

where $(J_T)_{ij} = \frac{\partial Y_i}{\partial T_{j}}$ and $(\mathbb{1})_{ij} = \delta_{ij}$.

A derivation of these derivatives is provided in appendix \ref{S-appendix}. We show additionally in appendix \ref{S-appendix_rel}, that iterative recurrent networks which are updated an infinite number of times lead, under certain conditions, to the same derivative.

In our simulations we have initialized the weights of $\vec{T}$ and $Q$ according to a uniform distribution in range of $[-0.5, 0.5]$.
The weights of $W$ were initialized to be all zero.

A implicit recurrent network is then trained according to the following pseudocode:

\begin{algorithm}
\caption{Training of implicit recurrent networks}\label{euclid}
\begin{algorithmic}[1]
\Procedure{Training step in epoch}{}
\State $D_{\vec{X}} \gets \text{load batch of input data}$
\State $D_{\vec{Y}_{gt}} \gets \text{load batch of ground truth data}$
\For {every input data $\vec{X}$ in $D_{\vec{X}}$}
\State Calculate $\vec{Y}$ so that equation \ref{eq:impl_function_methods} is full filled with accuracy $\epsilon$
\State Calculate derivatives of the implicit recurrent network
\State Compute and store gradient step for this input output pair
\EndFor
\State Perform a gradient step of the whole batch according to the ADAM algorithm \ref{sebsec:ADAM}
\EndProcedure
\end{algorithmic}
\label{pseudocode_one_layer_implicit_net}
\end{algorithm}

\subsection{Two-layer fully recurrent implicit networks}
\label{subsec:two-layer}

\begin{figure}
    \centering
    \includegraphics[width = 0.5\textwidth]{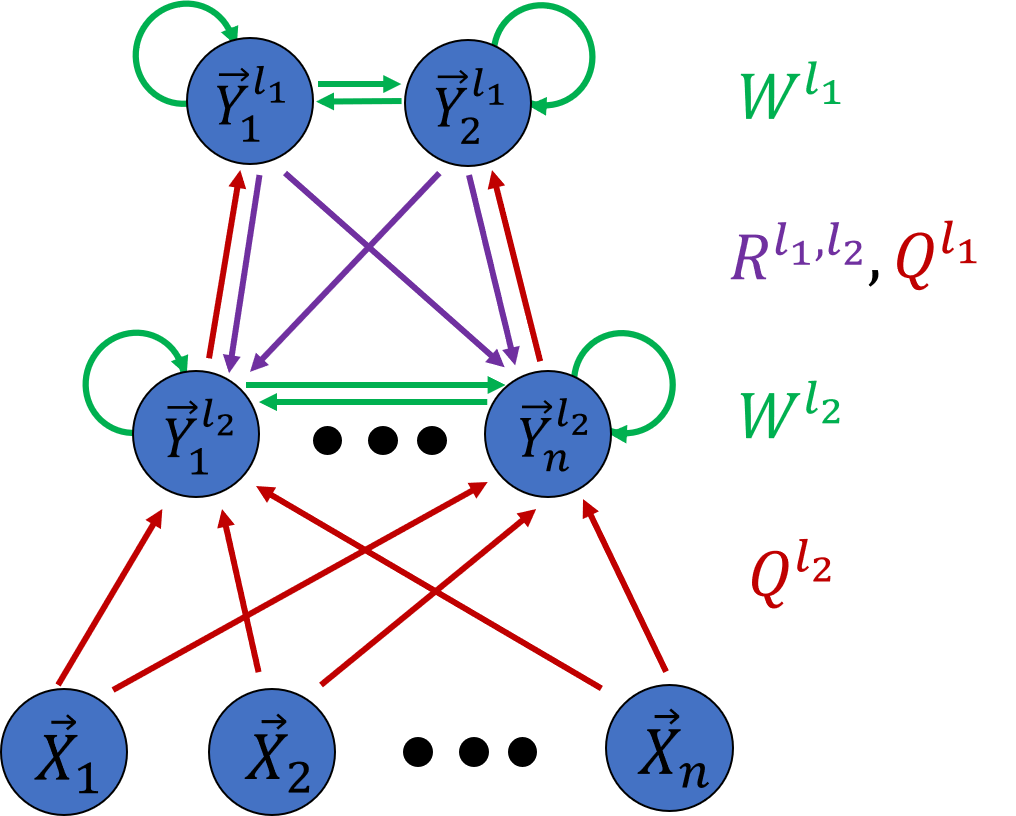}
    \caption{Depiction of a two-layer implicit recurrent network with lateral connections within layers and feed-back connections from the output into the hidden layer.}
    \label{fig:two_layer_impl}
\end{figure}

In this work, we also implement two-layer implicit recurrent layers, where additional feed-back connections from higher to lower layers are used.
An example of such a two-layer implicit recurrent network is depicted in figure \ref{fig:two_layer_impl}.

The activation of the output layer $l_1$ is then given by

\begin{equation*}
    \vec{Y}^{l_1} = f(Q^{l_1} \cdot \vec{Y}^{l_2} + W^{l_1} \cdot \vec{Y}^{l_1} + \vec{T}^{l_1}),
\end{equation*}

where $l_n$ denotes, to which layer the weights or the neural activation belongs to.

The activation of the second layer is then given by

\begin{equation*}
    \vec{Y}^{l_2} = f(Q^{l_2} \cdot \vec{X} + W^{l_2} \cdot \vec{Y}^{l_2} + R^{l_1 l_2} \cdot \vec{Y}^{l_1} + \vec{T}^{l_2}).
\end{equation*}

A derivation of the derivatives of this network is given in the appendix \ref{appen-two-layer}, where we show how to derive the derivative for the example of the derivative with respect to $W^{l_1}$.
In our experiments all recurrent weights $W^{l_1}$, $W^{l_2}$ and $R^{l_1 l_2}$ are initially set to zero. All other weights and biases are initialized uniformly within $[-0.5, 0.5[$.

\subsection{Semi-gradient implicit recurrent networks}
\label{subsec:semi-gradient}
In semi-gradient implicit recurrent networks, we ignore the reappearing dependence of $\vec{Y}$ with respect to the parameters.
This approximation is made, so that implicit recurrent networks can be implemented more easily into auto-differentiation modules such as pytorch \cite{paszke2019pytorch} or tensorflow \cite{abadi2016tensorflow}.
The procedure of training the semi-gradient implicit recurrent network remains in principle the same as described in the pseudo-code \ref{pseudocode_one_layer_implicit_net}, with the exception that neural activation which appear inside activation functions are treated as a constant.

\subsection{Calculation of the equilibrium}
\label{subsec:equi}

In order to use implicit activation functions, it is necessary that the equilibrium $\vec{Y}$ is calculated with a sufficiently low error $\epsilon$.
In our observation the iterative method of calculating the equilibrium $\vec{Y}$, does in principle work for small networks, but when many hidden neurons are used or the input dimension becomes too large, this method does not work at all.
Therefore, we compute the neural activation $\vec{Y}$ as proposed by P. Földiák \cite{foldiak1990forming} by solving the differential equation

 \begin{equation*}
     \frac{d \vec{Y}}{d t} = f( Q \cdot \vec{X} + W \cdot \vec{Y} + \vec{T}) - \vec{Y}
     \label{eq:dydt}
 \end{equation*}
 
numerically at equilibrium, where $\frac{d \vec{Y}}{dt} = 0$. 
 
If the Euler method is used in order to solve differential equation numerically, we obtain the update rule
 
 \begin{equation*}
    \vec{Y}(t+1) = \vec{Y}(t) + h \cdot (f(\widetilde{X}(t)) - \vec{Y}(t)).
  \end{equation*}
  
If a step size of $h = 1$ is used, this method is actually the same as the iterative method where $ \vec{Y}(t+1) = f(\widetilde{X}(t))$.
But, of course a step size of $h = 1$ is a very rough numerical approximation to the differential equation. This may explain, why the iterative method fails for big networks.
Therefore, we use from scipy.integrate.solve\_ivp \cite{virtanen2020scipy} the RK-4 method for 30 iterations in order to solve the equilibrium.
Here, we always start at $\vec{Y}(t=0) = 0$ and usually obtain an accuracy of $\lVert \frac{d \vec{Y}}{d t} \rVert \approx 10^{-8}$.
\\
In the case of two-layer implicit recurrent networks (see section \ref{subsec:two-layer}) the activation of both layers is coupled to one another.

Therefore, in order to obtain the equilibrium, we have to solve the differential equations

\begin{equation}
    \frac{ d \vec{Y}^{l_2}}{d t} = f(Q^{l_2} \cdot \vec{X} + W^{l_2} \cdot \vec{Y}^{l_2} + R^{l_1 l_2} \cdot \vec{Y}^{l_1} + \vec{T}^{l_2}) - \vec{Y}^{l_2}
\end{equation}

and 
\begin{equation}
    \frac{d \vec{Y}^{l_1}}{d t} = f(Q^{l_1} \cdot \vec{Y}^{l_2} + W^{l_1} \cdot \vec{Y}^{l_1} + \vec{T}^{l_1}) - \vec{Y}^{l_1}.
\end{equation}

 simultaneously.

\subsection{Learning rate optimizer}
\label{sebsec:ADAM}
We have used ADAM \cite{da2014method} with a learning rate of $0.01$ as an optimizer for our networks.
Here, we have applied the ADAM algorithm for every weight matrix $Q, W$  and $T$ separately.

\section{Discussion} %%%%%%%%%%%%%%%%%%%%%%%%%%%%%%%%%%%%%%%%
      \label{S-features}      

In this work, we show that a one-layer implicit recurrent neural network is able to solve the XOR-Problem. Therefore, this indicates that implicit recurrent neural networks are computationally more capable than state-of-the-art feed forward networks.
Furthermore, due to these results, we speculate that this indicates that the universal approximation theorem \cite{cybenko1989approximations} may hold for one-layer implicit recurrent neural networks.
The universal approximation theorem states, that a two-layer feed-forward network can approximate any real valued function, if it has enough hidden neurons.
Therefore, a two-layer feed-forward network can in principle solve any problem.
But in practise, it often works better to increase the number of layers in order to increase the computational power of the artificial neural network.
However, if too many layers are used within a network the problem of the vanishing/exploding gradient occurs, so that training of the weights within the first layers of the network becomes difficult.

Our work indicates that with the use of implicit recurrent neural networks, it is also possible to increase the computational power of neural networks. Thus, we think that implicit recurrent networks may be a new alternative to adding more neurons and more layers into artificial neural networks.
Therefore, we suggest to generalize our method to multi-layer recurrent neural networks. Thereafter, one could compare the network performance of feed-forward networks to implicit recurrent networks over the network depth.
We speculate, that implicit recurrent networks will reach accuracy's that can not be reached by a feed-forward network architecture.

Interestingly, in our simulations implicit recurrent networks show lower variance in the training error.
This suggests, that the network`s predictions are more consistent than in feed-architectures. 
Therefore, we suppose that it may be interesting to apply these networks to deep reinforcement learning algorithms, such as \cite{van2015deep, haarnoja2018soft}, which struggle with high variance during training.
Additionally, recurrent connections may make it possible to have a correction mechanism from higher to lower layers. Thus, we think it may be interesting to test weather actor-critic architectures with a shared layer architecture may benefit from such recurrent mechanisms.
Finally, we suggest to generalize implicit recurrent networks to convolutional networks \cite{lecun1995convolutional}, where recurrent connections between different filters could be added. Additionally, one could try to connect neighbouring kernels with one another.

\FloatBarrier
\bibliographystyle{unsrt}
\bibliography{literature}

\appendix

\section{Derivation of the derivatives of a one-layer implicit recurrent neuron} %%%%%%%%%
    \label{S-appendix}
    
In this section, we are going to derive the derivatives of a one-layer implicit recurrent neuron, which are necessary for the backpropagation update rule.
First, we are going to derive the derivatives directly from the implicit activation function.
Secondly, we will provide an alternative derivation, where we derive the derivative from an infinite chains of iterative network updates.

\subsection{Derivation from the implicit activation function}

In this section we consider a single neuron layer $\vec{Y}$, which receives a stationary input vector $\vec{X}$ over the feed-forward weight matrix $Q$ and has a bias vector $\vec{T}$. The neuron layer is additionally recurrently connected to itself over the recurrent weight matrix $W$. The overall input $\widetilde{X} = W \cdot \vec{Y} + Q \cdot \vec{X} + \vec{T}$ is then put into a sigmoid activation function $f(\cdot)$. \\
Therefore, the activation of this implicit recurrent network is given by

\begin{equation}
    \vec{Y} = f(W \cdot \vec{Y} + Q \cdot \vec{X} + \vec{T}).
    \label{eq:impl_function}
\end{equation}

If we now want to calculate the derivative of a specific neuron $Y_i$ in this layer with respect to a specific weight $W_{jm}$, we obtain

\begin{equation}
    \begin{split}
    \frac{\partial Y_i}{\partial W_{jm}} & = \frac{\partial f(\widetilde{X}_i)}{\partial \widetilde{X}_i} \cdot \frac{\partial \widetilde{X}_i}{\partial W_{jm}} = \frac{\partial f(\widetilde{X}_i)}{\partial \widetilde{X}_i} \cdot \left( \sum_k \left( W_{ik} \cdot \frac{\partial Y_k}{\partial W_{jm}} + \delta_{ij} \cdot \delta_{km} \cdot Y_k \right) \right) \\
     & = \frac{\partial f(\widetilde{X}_i)}{\partial \widetilde{X}_i} \cdot \left( \sum_k \left( W_{ik} \cdot \frac{\partial Y_k}{\partial W_{jm}} \right) + \delta_{ij}  \cdot Y_m  \right).
    \end{split}
     \label{eq:impl_deriv_w}
\end{equation}

In this equation the derivative of $\frac{\partial Y_i}{\partial W_{jm}}$ reappears on the right side of the equation. But, since in the sum over $k$ the derivatives of the other neurons $Y_k$ with respect to the weight $W_{jm}$ occurs as well, we obtain a set of $y \times y \times y$ linear equations.
This set of linear equations has to be solved in order to obtain the derivatives.

With the use of matrix formulation, equation \ref{eq:impl_deriv_w} can be rewritten to

\begin{equation*}
    J_W = f^{\prime} \odot \left(  W \cdot J_W + \delta_{\vec{Y}} \right),
\end{equation*}
    
where $(J_W)_{ijm} :=  \frac{\partial Y_i}{\partial W_{jm}}$, $(\delta_{\vec{Y}})_{ijm} := \delta_{ij}  \cdot Y_m $ and $(f^{\prime})_{i} := \frac{\partial f(\widetilde{X}_i)}{\partial \widetilde{X}_i}$.
Here, $\odot$ is simply the element wise multiplication sign between two matrices.

In order to obtain the resulting set of derivatives, one now has to solve  this equation for $J_W$, where one obtains

\begin{equation*}
     J_W = ( \mathbb{1} - f^{\prime} \odot W)^{-1} f^{\prime} \odot \delta_{\vec{Y}}.
\end{equation*}

Similarly, for the other derivatives with respect to $Q_{jm}$ one obtains 

\begin{equation*}
     J_Q = ( \mathbb{1} - f^{\prime} \odot W)^{-1} f^{\prime} \odot \delta_{\vec{X}},
\end{equation*}

where $(J_Q)_{ijm} = \frac{\partial Y_i}{\partial Q_{jm}}$ and $(\delta_{\vec{X}})_{ijm} = \delta_{ij}  \cdot X_m$.

For the derivatives with respect to $T_j$ one obtains

\begin{equation*}
     J_T = ( \mathbb{1} - f^{\prime} \odot W)^{-1} f^{\prime} \odot  \mathbb{1} , 
\end{equation*}

where $(J_T)_{ij} = \frac{\partial Y_i}{\partial T_{j}}$ and $(\mathbb{1})_{ij} = \delta_{ij} $.

Therefore, we have derived all derivatives, which are necessary for applying the backpropagation algorithm in a one-layer implicit recurrent neural network.

\subsection{Derivation as an infinite chain of iterative recurrent network updates}
\label{S-appendix_rel}

Alternatively, these derivatives can also be derived by considering an iterative recurrent neural network, which uses infinite iterative update steps.
In this derivation, we additionally assume that the iterative update reaches an equilibrium after a finite iterative updates. Furthermore, we assume that $(f^{\prime} \odot W)^{k} \approx 0$ for large $k$, where $f^{\prime}$ is the derivative of the activation function at the equilibrium.
This assumption is analogous to the assumption, that the gradient vanishes after many iterative updates.

In order to show that let us consider the iterative update rule for an iterative one-layer recurrent neural network.

The activation function is then given by:

\begin{equation}
    \vec{Y}(t) = f(W \cdot \vec{Y}(t-1) + Q \cdot \vec{X} + \vec{T})
    \label{eq:iter_function}
\end{equation}

In the firs time step we set $\vec{Y}(t = 0) = 0$ and this activation function is then updated iteratively up to $\vec{Y}(t = n)$.

If we calculate for example the derivative with respect to $W_{jm}$, we obtain

\begin{equation}
    \begin{split}
    \frac{\partial Y_i (t)}{\partial W_{jm}}  = \frac{\partial f(\widetilde{X}_i)}{\partial \widetilde{X}_i} \cdot \left( \sum_k \left( W_{ik} \cdot \frac{\partial Y_k (t-1)}{\partial W_{jm}} \right) + \delta_{ij}  \cdot Y_m (t-1)  \right),
    \end{split}
     \label{eq:iter_deriv_w}
\end{equation}

which can again be rewritten into matrix formulation by

\begin{equation*}
    J_W(t) = f^{\prime}_{\widetilde{X}(t-1)} \odot \left(  W \cdot J_W(t-1) + \delta_{\vec{Y}(t-1)} \right),
\end{equation*}

where $f^{\prime}_{\widetilde{X}(t-1)}$ is the derivative of the activation function for the overall input at the $(t-1)$ th iterative time step.
This calculation has to be applied repeatedly up to the first time step, therefore we obtain for $n >= 2$ 

\begin{equation}
    J_W(t) = \sum_{k= 0}^{n} \left[ \left ( \prod_{m = 1}^{k} f^{\prime}_{\widetilde{X}_{(t-m)}} \odot W \right) \cdot f^{\prime}_{\widetilde{X}_{(t-k-1)}} \odot \delta_{\vec{Y}(t -k - 1)} \right).
    \label{eq:iter_deriv_w_derivation}
\end{equation}

Because we apply an infinite amount of iterative time steps $n$ goes to infinity.
Furthermore, we assume that this system reaches an equilibrium, where $\vec{Y}(t) = \vec{Y}(t-1)$, after a finite iterative update steps.

Therefore, we can approximate equation \ref{eq:iter_deriv_w_derivation} to

\begin{equation*}
    J_W(t) \approx   \sum_{k= 0}^{\infty} \left ( \prod_{m=1}^{k} f^{\prime} \odot W \right) \cdot f^{\prime} \odot \delta_{\vec{Y}(t)}  = \left(  \sum_{k= 0}^{\infty} (f^{\prime} \odot W)^{k} \right) \cdot f^{\prime} \odot \delta_{\vec{Y}(t)} ,
\end{equation*}

where we have used that the last infinite iterative updates are calculated at the equilibrium. Thus, we can approximate $\delta_{\vec{Y}(t - k - 1)}$ for these time steps to the equilibrium $\delta_{\vec{Y}(t)}$. Additionally, we can therefore approximate $f^{\prime}_{\widetilde{X}(t-m)}$ to $f^{\prime}$ which is the derivative at equilibrium.

Next, in order to further simplify this equation, we can use the Neumann series, which is a generalisation of the geometric series to matrices.
Therefore, we additionally assume that $\sum_{k= 0}^{\infty} (f^{\prime} \odot W)^{k}$ is convergent, from which follows that $\sum_{k= 0}^{\infty} (f^{\prime} \odot W)^{k} = ( \mathbb{1} - f^{\prime} \odot W)^{-1}$.

We can now use this fact to obtain the final iterative update rule after a infinite amount of update rules, which is then given by

\begin{equation*}
     J_W = ( \mathbb{1} - f^{\prime} \odot W)^{-1} f^{\prime} \odot \delta_{\vec{Y}}.
\end{equation*}

This final derivative with respect to $W$ is the same as the derivative for the implicit recurrent network and for the other derivatives with respect to $Q$ and $T$ this can be shown analogously.
Therefore the relationship between the derivative of an implicit recurrent network and the derivative of an iterative recurrent network, which is updated an infinite number of times at the equilibrium, has been shown.

\section{Derivation of derivatives of two-layer implicit recurrent neural networks}
\label{appen-two-layer}

In this section the derivatives of two-layer implicit recurrent neural networks will be derived.
Here, the output layer $\vec{Y}^{l_1}$ is defined as 

\begin{equation*}
    \vec{Y}^{l_1} = f(Q^{l_1} \cdot \vec{Y}^{l_2} + W^{l_1} \cdot \vec{Y}^{l_1} + \vec{T}^{l_1}),
\end{equation*}

where $l_n$ denotes, to which layer the weights or the output belongs to.

The activation of the second layer is then given by

\begin{equation*}
    \vec{Y}^{l_2} = f(Q^{l_2} \cdot \vec{X} + W^{l_2} \cdot \vec{Y}^{l_2} + R^{l_1 l_2} \cdot \vec{Y}^{l_1} + \vec{T}^{l_2}),
\end{equation*}

where $R^{l_1 l_2}$ are feedback connections from layer $l_1$ to layer $l_2$.

If we calculate the derivative of both layers with respect to $W^{l_1}_{ij}$ and write them down into matrix notation, we obtain for the first layer $\vec{Y}^{l_1}$

\begin{equation}
    J^{l_1}_{W^{l_1}} = f^{\prime}_{l_1} \odot ( Q^{l_1} \cdot J^{l_2}_{W^{l_1}} + \delta_{\vec{Y}^{l_1}} + W^{l_1} \cdot J^{l_1}_{W^{l_1}}),
    \label{eq:dy_l1}
\end{equation}

where $(J^{l_1}_{W^{l_1}})_{ijm} = \frac{\partial Y^{l_1}_i}{\partial W^{l_1}_{jm}}$, $(J^{l_2}_{W^{l_1}})_{ijm} = \frac{\partial Y^{l_2}_i}{\partial W^{l_1}_{jm}}$, $(\delta_{\vec{Y}^{l_1}})_{ijm} := \delta_{ij}  \cdot Y^{l_1}_m $ and $f^{\prime}_{l_1}$ is simply the derivative of the sigmoid activation functions in layer $l_1$.

The derivatives for layer $\vec{Y}^{l_2}$ are analogously given by

\begin{equation}
    J^{l_2}_{W^{l_1}} = f^{\prime}_{l_2} \odot ( R^{l_1 l_2} \cdot J^{l_1}_{W^{l_1}}+ W^{l_2} \cdot J^{l_2}_{W^{l_1}}).
    \label{eq:dy_l2}
\end{equation}

Therefore, we obtain with equation \ref{eq:dy_l1} and \ref{eq:dy_l2} two coupled sets of linear equations, which have to be solved in order to obtain $J^{l_1}_{W^{l_1}}$ which can then be used for backpropagation.
The first step would be to solve equation \ref{eq:dy_l2} for $J^{l_2}_{W^{l_1}}$.
Here, we obtain

\begin{equation*}
    J^{l_2}_{W^{l_1}} = (\mathbb{1} - f^{\prime}_{l_2} \odot W^{l_2})^{-1} \cdot f^{\prime}_{l_2} \odot ( R^{l_1 l_2} \cdot J^{l_1}_{W^{l_1}}).
    \label{eq:dy_l2_solved}
\end{equation*}

Then this solution can be used by inserting the solution into equation \ref{eq:dy_l1}, where we again have to solve for $J^{l_1}_{W^{l_1}}$.

The final solution is then given by 

\begin{equation*}
    J^{l_1}_{W^{l_1}} = (\mathbb{1} - f^{\prime}_{l_1} \odot W^{l_1} - f^{\prime}_{l_1} \odot Q^{l_1} \cdot (\mathbb{1} - f^{\prime}_{l_2} \odot W^{l_2})^{-1} f^{\prime}_{l_2} \odot R^{l_1 l_2})^{-1} \cdot f^{\prime}_{l_1} \odot \delta_{\vec{Y}^l_1}.
\end{equation*}

Note that, if we set $R^{l_1 l_2} = 0$ the derivatives simplify to
\begin{equation*}
    J^{l_1}_{W^{l_1}} = (\mathbb{1} - f^{\prime}_{l_1} \odot W^{l_1})^{-1} \cdot f^{\prime}_{l_1} \odot \delta_{\vec{Y}^l_1},
\end{equation*}
which has the same form as the derivatives of a one-layer implicit recurrent network.

The other derivatives with respect to $T_{l_1}$, $Q_{l_1}$, $T_{l_2}$, $R_{l_1 l_2}$, $W_{l_2}$ and $Q_{l_2}$ can be derived analogously.

The derivatives are then given by

\begin{equation*}
    J^{l_1}_{Q^{l_1}} = S  \cdot f^{\prime}_{l_1} \odot \delta_{\vec{Y}^l_2},
\end{equation*}

\begin{equation*}
    J^{l_1}_{T^{l_1}} = S \cdot f^{\prime}_{l_1} \odot \mathbb{1},
\end{equation*}

\begin{equation*}
    J^{l_1}_{Q^{l_2}} = S \cdot f^{\prime}_{l_1} \odot Q^{l_1} (\mathbb{1} - f^{\prime}_{l_2} \odot W^{l_2})^{-1} \cdot f^{\prime}_{l_2} \odot \delta_{\vec{X}}.
\end{equation*}

\begin{equation*}
    J^{l_1}_{R^{l_1 l_2}} = S \cdot f^{\prime}_{l_1} \odot Q^{l_1} (\mathbb{1} - f^{\prime}_{l_2} \odot W^{l_2})^{-1} \cdot f^{\prime}_{l_2} \odot \delta_{\vec{Y}^{l_1}},
\end{equation*}

\begin{equation*}
    J^{l_1}_{W^{l_2}} = S \cdot f^{\prime}_{l_1} \odot Q^{l_1} (\mathbb{1} - f^{\prime}_{l_2} \odot W^{l_2})^{-1} \cdot f^{\prime}_{l_2} \odot \delta_{\vec{Y}^{l_2}}
\end{equation*}

and 

\begin{equation*}
    J^{l_1}_{T^{l_2}} = S \cdot f^{\prime}_{l_1} \odot Q^{l_1} (\mathbb{1} - f^{\prime}_{l_2} \odot W^{l_2})^{-1} \cdot f^{\prime}_{l_2} \odot \mathbb{1},
\end{equation*}

where $S := (\mathbb{1} - f^{\prime}_{l_1} \odot W^{l_1} - f^{\prime}_{l_1} \odot Q^{l_1} \cdot (\mathbb{1} - f^{\prime}_{l_2} \odot W^{l_2})^{-1} f^{\prime}_{l_2} \odot R^{l_1 l_2})^{-1}$.

\end{document}